\documentclass[runningheads]{llncs}
\usepackage{graphicx}
\usepackage{cite}

\usepackage[linesnumbered,ruled,lined]{algorithm2e}
\usepackage{algpseudocode}
\usepackage{booktabs}

\usepackage{tabularx}
\usepackage{array}
\usepackage{multirow, makecell}
\usepackage{multicol}
\usepackage{arydshln}
\usepackage{pifont}
\usepackage{xcolor}
\usepackage{color,soul}

\begin{document}
\title{Federated Learning: \\ Issues in Medical Application}

\author{Joo Hun Yoo\and Hyejun Jeong \and
Jaehyeok Lee \and \\ Tai-Myoung Chung*}

\authorrunning{J.H.Yoo, H.Jeong et al.}

\institute{College of Computing and Informatics, Sungkyunkwan University, \\ Suwon, Korea (Republic of)\\
\email{\{andrewyoo, june.jeong, esinfam99\}@g.skku.edu}\\
\email{\{tmchung\}@skku.edu}}
\maketitle              

\begin{abstract}

Since the federated learning, which makes AI learning possible without moving local data around, was introduced by google in 2017 it has been actively studied particularly in the field of medicine. In fact, the idea of machine learning in AI without collecting data from local clients is very attractive because data remain in local sites. However, federated learning techniques still have various open issues due to its own characteristics such as non identical distribution, client participation management, and vulnerable environments. In this presentation, the current issues to make federated learning flawlessly useful in the real world will be briefly overviewed. They are related to data/system heterogeneity, client management, traceability, and security. Also, we introduce the modularized federated learning framework, we currently develop, to experiment various techniques and protocols to find solutions for aforementioned issues. The framework will be open to public after development completes.

\end{abstract}
\section{Introduction}

Since the advent of Artificial Intelligence-based learning techniques, machine learning, have been extensively studied in the medical field such as radiology, pathology, neuroscience, and genetics. Instead, machine learning or deep learning techniques identify hidden multi-dimensional patterns to predict results, which come as a huge gain in disease diagnosis that requires a higher understanding of nearly human-unidentifiable correlation. However, most medical machine learning researches often ignore data privacy.

Federated learning has emerged in reaction to the strengthened data regulations on personal data such as California’s Consumer Privacy Act (CCPA), EU’s General Data Protection Regulation (GDPR), and China’s Cyber Security Law (CSL). Those regulations not only restrict the reckless use of personal information but prevent private data collection. Thus, centralized methods, which collect and learn based on enormous amount of data, is constrained under the regulations. Centralized learning techniques are even more difficult to be applied in the field of medicine, where data contains a lot of sensitive information. In particular, Personal Health Information (PHI) contains a number of sensitive personal information such as names, addresses, and phone numbers, so collecting and using them for the learning process is against the worldwide privacy acts. 

While primary information leakage can be prevented through various security methods such as pseudonymization, secure aggregation, data reduction, data suppression, and data masking, other variables like images or biomarkers can also identify data owners. Thus, it cannot said to be a complete solution for data privacy. 

Federated learning has emerged for stronger data protection than the aforementioned de-identification methods. It has unique characteristics compared to centralized machine learning. Specifically, traditional machine learning requires a large volume of training data; it mandates the collection of personal data to a centralized server. Decentralized federated learning, however, generates and develops deep neural network models without data collection. It consequently allows researchers to apply machine learning methods without ruining data ownership. The most widely used federated learning framework is accomplished through repeating four steps, which will be discussed in the next section.

The structural advantages of federated learning are a huge gain in medical field. Data held by each client can be used as training data without leaving its local position. It results in utilizing more data while protecting highly sensitive medical data. In addition, the application of federated learning techniques also allows the participating hospitals to connect so as to make more accurate and generalized models. However, it is difficult to deploy the current structure due to several open issues such as data vulnerability and model poisoning as mentioned in \cite{qayyum2020secure}.

This paper is organized as follows: in the next section, we describe how federated learning is currently applied to the medical field with specific examples. In Section three, we summarised the problems arising from the application of federated learning in healthcare and provide solutions in the following order: data and system heterogeneity, client management, traceability and accountability, and security and privacy.

\section{Federated Learning in Medical Applications}

\begin{figure}[h]
\centering
\includegraphics[width=\textwidth]{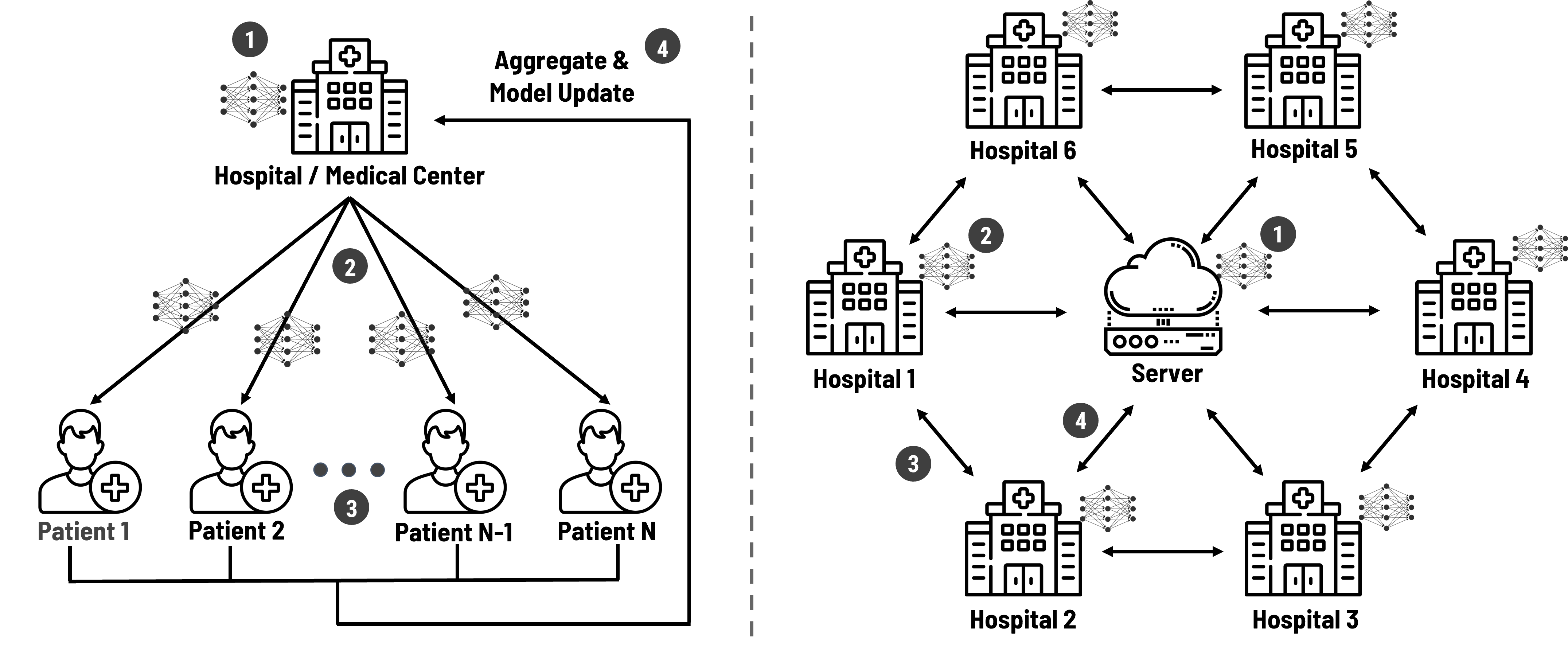}
\caption{Federated learning architecture in the field of medicine.}
\label{fig:medicalFL}
\end{figure}

Federated learning, which enables the application of various machine learning methods without data collection in distributed environments, is recently widely used in healthcare. Figure \ref{fig:medicalFL} illustrates two representative architectures of medical federated learning. The left figure is when patients are the local clients and the hospital is the central server, and the right one is when participating clients are the hospitals with a server at the center.

A server transmits its initial model to each client, a hospital or a patient, for example. Each local client trains the received model, and then sends only model parameters back to the server. The server then aggregates all the received model parameters to update the global model; consequently this collaborative and distributed learning can have the same effect of centralized learning. 
Most importantly, the global model is updated via aggregating multiple local models, so the local data privacy is isolated. 

The use of medical data has become heavily available with the development of federated learning. Medical data is especially difficult to handle because it is normally in large volume, and it is restricted under data regulations. Federated learning, however, allows to utilize these data as it does not involve raw data transmission.

\subsection{Physical Disorder Predictions}

Deep learning techniques are used in many studies in medical field, such as finding lesions and predicting patients' disorders. With aid of federated learning, recent studies focuses not only to improve prediction accuracy but data privacy.

Liu \cite{liu2020experiments} and Zhang\cite{zhang2021dynamic} conducted experiment on diagnosing COVID-19 in both federated and centralized learning, demonstrating that both prediction accuracy are comparable.
Choudhury conducted experiments to predict Adverse Drug Reactions with Electronic Health Records dataset \cite{choudhury2019predicting} and compared the results of that used centralized learning, localized learning, and federated learning. Even Baheti's experiment showed higher accuracy in pulmonary lung nodules detection by applying federated learning \cite{baheti2020federated}.

Roth\cite{roth2020federated} and Jimenez\cite{jimenez2021memory} used mammography dataset to classify breast density and cancer, respectively. Both results showed higher classification accuracy in federated learning than centralized learning. Yi proposed SU-Net for brain tumor segmentation, which also showed a higher AUC in federated environment \cite{yi2020net}.

\subsection{Mental Disorder Predictions}
With the development of artificial intelligence, medical data is available to predict mental illness, as well as physical illness.
Xu proposed FedMood to analyze and diagnose depression \cite{xu2021fedmood}. FedMood deployed DeepMood \cite{suhara2017deepmood} architecture with the data collected from multiple smartphones while typing on the mobile phone and the user's HDRS score to analyze subjects' psychological states.
Yoo proposed PFCM that implements a hierarchical clustering method to classify the severity of major depressive disorder using heart rate variability \cite{yoo2021personalized}. 
Liu proposed DTbot that is designed for depression treatment \cite{liu2021federated}. DTbot collects user's voices and video in real-time and analyzes emotions. If pessimistic emotions are recognized, it plays music that fits the users' taste to relax their mood.\\

\noindent
Numerous recent works applied federated learning to medical data. Their prediction accuracy is higher or comparable to that of centralized machine learning. It consequently suggests the potential of federated learning, not only performance enhancement but data privacy preservation, so as to let practitioners can use it for further uses such as disease prediction. 

\subsection{Other Medical Applications}
Along with the two aforementioned prediction tasks, federated learning is also applied for health monitoring using smart devices or life signal monitoring.
 
Chen \cite{chen2020fedhealth} and Wu\cite{wu2020fedhome} used federated learning for activity recognition. 
Chen proposed FedHealth, a federated transfer learning framework for wearable devices, to build a more personalized model \cite{chen2020fedhealth}. 
Wu proposed FedHome framework and GCAE method to provide a more personalized and accurate health monitoring for in-home elders \cite{wu2020fedhome}. 
Can analyzed cardiac activity data collected from smart bands for stress-level monitoring on various occasions \cite{can2021privacy}.
Yuan proposed a federated learning framework for healthcare IoT devices that relaxes the computation load on them and communication overhead between the edge devices and a server \cite{yuan2020federated}.  

All their proposed methods outperformed centralized methods. It indicates that federated learning can be used in a variety, such as disease management and prevention, addiction or mental health tracking, and real-time health monitoring \cite{hakak2020framework}.

\section{Research Issues}

\subsection{Heterogeneity Issues}

Although federated learning demonstrates strong performance in applying machine learning techniques while adhering to data privacy regulations, most of the works rely on an Independent and Identically Distributed (IID) assumption. However, in most real-world data environments, including the field of medicine, this assumption is not followed. We classify these non-IID situations into data heterogeneity and system heterogeneity in federated learning.

\subsubsection{Data Heterogeneity}

Data heterogeneity, called non-IID data distribution, means that data held by participating clients in federated learning are not uniformly distributed but are in heterogeneous distribution or characteristics. A non-IIDness can be decoupled into not identical and not independent distribution. Non-identical client distributions can be categorized into five: a feature distribution skew, label distribution skew, same label but different features, same features but different labels, and quantity skew \cite{kairouz2019advances}. We give practical examples of the five non-IID cases in a federated learning environment, especially in the medical field.

\begin{itemize}
    \item \emph{Feature distribution skew}: Even if two individuals wear the same smartwatch model and exercise for the same time duration, the features of measured values are unique because each person has different characteristics such as posture and heart rate.\\
    
    \item \emph{Label distribution skew}: Frostbite is a disease that frequently occurs in cold areas because it is caused by exposure to severe cold resulting in tissue damage to body parts. Therefore, it is rare in places with relatively warm temperatures. \\

    \item \emph{Same label but different features}: Professional medical devices are used to measure healthcare data such as neuroimages and biomarkers of patients. Medical devices, however, are made by distinct brands chosen by the different hospitals. \\
    
    \item \emph{Same features but different labels}: Lung photographs damaged by the recent global pandemic COVID-19 virus are difficult to distinguish from those of pneumonia because they have common characteristics in many lesions. \\

    \item \emph{Quantity skew}: Suppose five times more patients have visited hospital \textit{A} than hospital \textit{B}. The quantity of data each hospital has will also significantly differ.

\end{itemize}

Not-independent distribution is a violation of consistency of data depending on the other factors. Such violations are introduced when the data changes over time or geolocation \cite{kairouz2019advances}. 

\subsubsection{System Heterogeneity}

Participating devices may cause system heterogeneity depending on their hardware setting, computing power, communication cost, and network connectivity \cite{li2020federated}. Considering the federated learning environment where multiple medical centers participate in the learning process, it can cause the differences in database or infrastructure between each hospital. For example, Samsung Medical Center has built a systematic infrastructure for efficient management of patient data through operating their own database and digital therapeutics research center. On the other hand, most clinics are more not likely to be equipped with such systems. These differences raise system heterogeneity issues when they participate in federated learning.

\subsubsection{Approaches for Heterogeneity Issues}
Data and system heterogeneity issues have various misleading consequences in the federated learning environment. As the distribution of each data varies, the global model does not converge to a single global model; as clients have different capabilities, the global model may be biased into a dimension. Numerous researches have been conducted to address the issue, and we group the works into the three main approaches: clustering, optimization, and model fusion.

\subsubsection*{Clustering methods}

Machine learning clustering-based techniques are the representative unsupervised learning methods for finding similarities with peripheral data for unlabeled datasets. Many researchers leveraged clustering methods to solve data heterogeneity issues because gathering data points with similar patterns from unlabeled data is analogous to grouping clients with similar weight distribution to which the server is inaccessible in federated learning.

Sattler proposed CFL that adopts a clustering method by measuring the cosine similarity of each local model \cite{sattler2020clustered}. They swapped the labels to fit the dataset for the non-IID environment. Their experiment results demonstrated that their proposed work achieved a reasonable performance even in extreme non-IID situations.
Briggs applied a hierarchical clustering-based method to improve the performance when the clients have non-IID data \cite{briggs2020federated}. They introduced a method to generate optimized clusters by comparing L1, L2, and cosine similarity distances between each cluster, demonstrating that the proposed method can reach the desired performance faster and more accurately than traditional federated learning methods. 
Based on \cite{briggs2020federated}, Yoo used Heart Rate Variability data of patients to diagnose depression severity \cite{yoo2021personalized}. They applied clustering-based algorithms called PFCM for new incoming participants to improve prediction accuracy and solve non-IID issues.

While there are various clustering-based techniques to deal with the performance degradation caused by data heterogeneity, there are also researches to solve the issues of model aggregation time delay caused by heterogeneous data. Chen introduced the FedCluster method to address the problem of slow convergence that occurs when the federated averaging method is applied to heterogeneous local data \cite{chen2020fedcluster}. Each participating local client is not included in the model update at a time, but is clustered according to certain criteria, in which the cluster participates in the federated learning process for each round. Depending on the federated learning environment they want to apply, they apply the best clustering scenario of random uniform clustering, timezone-based clustering, and availability-based clustering. Experiments with MNIST and CIFAR-10 benchmark datasets demonstrated that the cyclic federated learning structure through FedCluster showed a faster convergence time than the conventional FedAvg algorithm.

As a server has no knowledge in the distribution of the participants' data, a number of works tried to address this issue from the perspective of unsupervised learning. Thus, clustering, one of the most prevalent unsupervised methods, is the most widely applied to solve the heterogeneity issues by gathering clients.

\subsubsection*{Optimization methods}
Another branch of solving the heterogeneity issue is the use of optimization algorithms. With the only use of FedAvg limits the models from converging as a single global model cannot properly represent non-IID data. This is because the data derived from different distributions diverge in various direction to represent the features of the distributions to which they belong.

Xie proposed federated SEM, a multi-center federated learning framework, which allows optimization function to find multiple local optima points \cite{xie2020multi}. The Stochastic Expectation Maximization makes it possible to find local optima point in a variety of distributions of clients in each clustered multi-center, which solves the problems that do not converge to a single global model. Reddi applied various adaptive optimizers called FEDADAGRAD, FEDYOGI, and FEDADAM, which are advantageous in hyperparameter coordination and improving the convergence rate of the federated learning model over the vanilla FedAvg method \cite{reddi2020adaptive}. 
SCAFFOLD also addresses client-drifting issues in federated learning. By introducing a client control variable, Karimireddy adjust each local update in the direction of the optimal global model \cite{karimireddy2020scaffold}.
FedProx from \cite{li2018federated} modifies the conventional FedAvg method in two directions; tolerating partial work and adding proximal terms. Since each participating device may differ in computing power or network bandwidth, Li solves the heterogeneity issues by assigning each device for a specific iteration. A proximal term is also used to prevent heterogeneity problems arising from excessive local update iterations.

\subsubsection*{Mixture model of global and local}

In addition to clustering and optimization based techniques, Hanzely \cite{hanzely2020federated} and Arivazhagan \cite{arivazhagan2019federated} solved heterogeneity issues by mixing global and local models generated by federated learning. Since a global model is too general to fit all clients’ data and local models are too specific to be generalized, they extract representative learning layers from both global and local to create a fused model. In this way, it is possible to adopt general features from other participating clients, while adding their personalized features from own local data.

\subsection{Client Management Issues}

Unlike the centralized machine learning architecture, client management is another essential issue in federated learning. The server must decide which client should participate in the learning process because some free-riding clients might be looking for benefits without any contribution.

In line with this, many researchers brought incentive mechanisms to federated learning. Incentive mechanism was first introduced by system architects aiming for improved performance of repetitive loops of manufacturers by rewarding the participants \cite{mitra2008competition}. Although many pieces of studies have been conducted so far, they had not utilized a proper incentive mechanism in federated learning because it is challenging to evaluate which participating client contributed how much.

\begin{figure}[h]
\centering
\includegraphics[width=\textwidth]{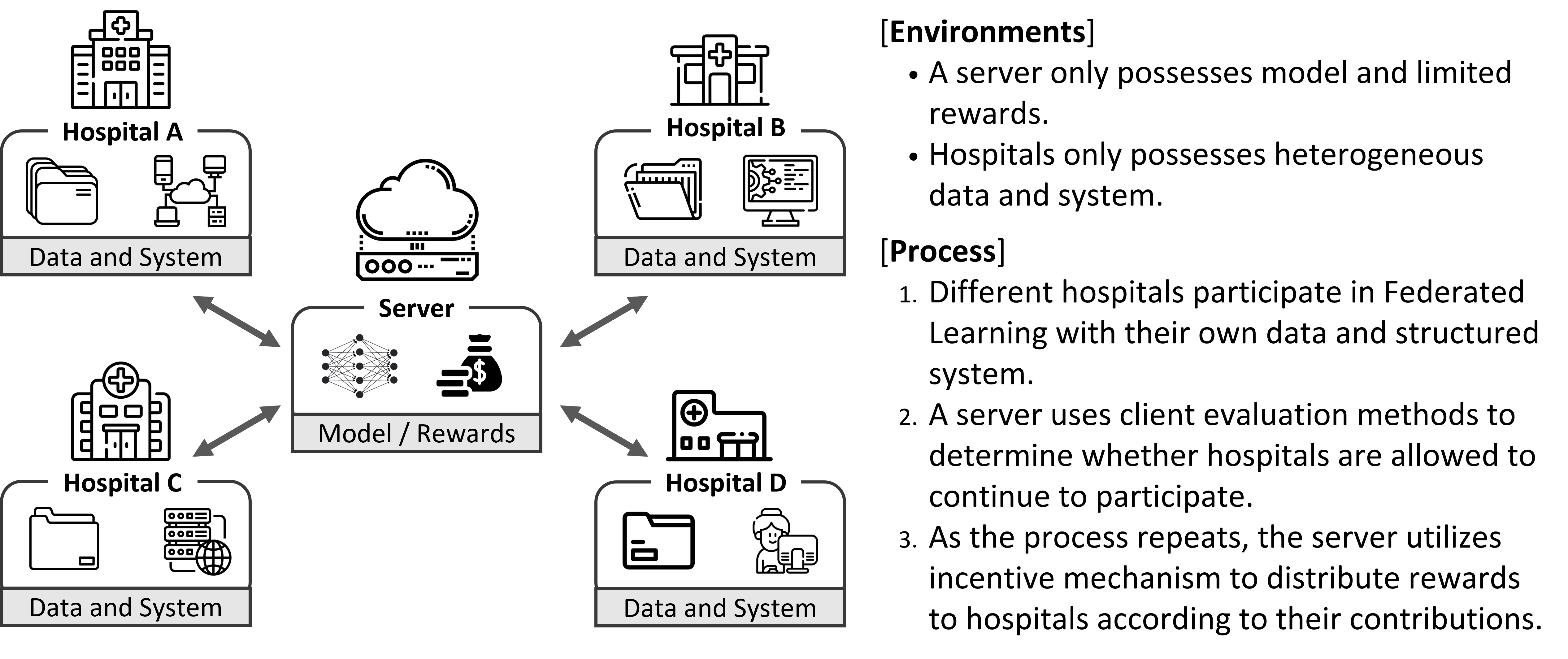}
\caption{Client Management in the field of medicine.}
\label{fig: clientman}
\end{figure}

\subsubsection{Client Management based on Data Quality}

Under strong data regulations, the high quality data becomes a large asset, so the local clients may want to benefit from the participation in federated learning systems. Thus, depending on the data quantity and quality, an incentive mechanism has emerged to reward participating local clients.

Chen deals with the incentive issue that occurs when multi-parties participate in federated learning to generate a better performing model collaboratively \cite{chen2020mechanism}. From an economic perspective, they pointed out that if other companies can grow further through the high-quality data of one good client, the client with high-quality data will not participate in federated learning because it will threaten their own profits.

\subsubsection{Client Management based on Resource Usage}

In addition to the importance of client data, it is critical to identify the computational capabilities or systems they use to participate in federated learning. Zeng examined a possible multi-dimensional resource differences that occur in Mobile Edge Computing (MEC) environments \cite{zeng2020fmore}. Servers must recognize dedicated resources and provide appropriate incentives to encourage good clients to participate in order to ultimately improve model performance. Most federated learning researches, on the other hand, assumed that each client participating in model training has the same dedication.

\subsubsection{Approaches for Client Management Issues}

Generating a better performance learning model requires contributions from multiple data providers with proper quality of data and resource, though not all clients equally contribute to federated learning. Therefore, an algorithm that can identify each contribution needs to be applied. The two most widely studied approaches leveraged Shapley value and Stackelberg game theory which we will discuss in the following.

\subsubsection*{Shapley Value}

Jia suggested how to perform data valuations through Shapley Value \cite{jia2019towards}, which has been widely used in game theory. They listed how Shapley Value enables data evaluation in multiple machine learning analytics environments and demonstrates their approach's scalability. Similar to the Shapley Value based game-theory approach, Lim used contract theory to identify the data quality and quantity of each data owner and applied a hierarchical incentive mechanism in the federated crowd-sourcing network \cite{lim2020hierarchical}.

\subsubsection*{Stackelberg Game Theory}

The Stackelberg game theory is widely used to assess each participant's contribution and construct a reward system. Sarikaya solved the problem caused by heterogeneous worker performance through the Stackelberg game-based method \cite{sarikaya2019motivating}. This measures the time it takes each participant to complete a given task for an updated gradient transfer and assigns a proper reward for each computing power based on the Stackelberg game theory.
Khan also adopted a Stackelberg game-based approach in \cite{khan2020federated}. Each edge node delivers its own computation energy and latency to the model aggregator, which is in charge of incentives. The goal of the model aggregator is to minimize learning time while maximizing model performance, so it adjusts client learning level based on the clients’ Stackelberg results. 

Pandey proposed a two-stage of Stackelberg game by developing an optimal learning model through maximizing the utility of participating devices and MEC servers \cite{pandey2020crowdsourcing}. When the MEC server announces the objectives of the optimized global model it wants to create and rewards accordingly, each device participates in the federated learning by optimizing the global learning model and maximizing the yield through the local data it possesses.

Similarly to studies that deal with data values by Stackelberg game theory, auction systems were also applied to solve clients’ heterogeneous resource issues. To address the problem of client management, Le used a primary-dual greedy auction mechanism \cite{le2021incentive}. When the server is assigned the task of federated learning training, each client submits a bid based on their own computation resource and transmission power. Subsequently, the server selects clients who can develop optimal models based on the bid list and provides customized rewards after completing the learning task. \cite{zeng2020fmore} also applied auction-based techniques on various scoring functions to allow devices with high-quality data to participate at a relatively low cost.\\

\noindent
Giving no or the same level of incentive to all local clients will result in some participants earning rewards for providing their low-quality data and resource. Others, on the other hand, will suffer from losses while contributing high-quality data. Hence, designing federated learning without explicit incentive mechanisms may violate the purpose of federated learning, which is to collaboratively develop a high-performance learning model. Researchers must develop a more sophisticated incentive mechanism to manage local clients in a real-world federated learning environment systematically.

\subsection{Traceability and Accountability Issues}

The most notable advantage of federated learning is data privacy, as the global server cannot directly investigate the local clients' data. However, this presents a substantial difficulty in that it is impossible to track results or hold them accountable for learning. Rieke highlighted a method for determining who is responsible for the unexpectedly erroneous outcomes of the medical analysis caused by the federated learning algorithm \cite{rieke2020future}. Indeed, the inability to check the learning process of AI is a problem that has arisen as application of deep learning techniques has expanded. This is due to the black-box nature of the neural network, and federated learning should consider taking another step toward data privacy.

Crucial factor in the field of medicine is the explainability of what the results were based on when making a decision. That is why precision and recall are more commonly used for performance evaluation than accuracy in the medical field. Results derived from the machine learning model trained with medical data should be reviewed more carefully, though it is hardly achievable in a federated learning environment.

It would be ideal if the predictive or diagnostic model obtained by the federated learning consistently demonstrated professional-level performance, but this is not the case. Therefore, researchers should consider when it produces false-positive results. If the false-positive rate is high in the federated learning task, which participant or training round is responsible for the problem should be determined. Otherwise, federated learning cannot find out which part is causing the issue; the entire training architecture may need to be redesigned.

While medical experts can provide sufficient information during data preprocessing, such as labeling, segmentation, noise filtering, this cannot happen during the federated learning's training process \cite{xu2021federated}. As a result, federated learning can be seen as a trade-off architecture in the medical field. Although it is critical to protect PHI from arbitrary inspection, the system's most critical advantage is used as a fatal disadvantage to healthcare applications. In line with this, Explainable AI (XAI) was developed to solve the black-box problem \cite{lundberg2017unified}, which allowed researchers to find out which parts of the deep neural networks are responsible for the performance degradation. These characteristics of XAI must also be applied when federated learning is adopted in the field of medicine to prevent medical errors caused by false-positive rates.

There have not been many studies that use XAI in the field of medical federated learning, but some researchers attempted to combine XAI and FL to increase explainability while maintaining the benefits of data privacy protections. Raza applied XAI with Federated Transfer learning to design ECG (Electrocardiography) monitoring healthcare system in \cite{raza2021designing}, by adding Gradient-weighted Class Activation Mapping (Grad-CAM) module on federated learning architecture to provide ECG signal classification task. However, more extensive researches on XAI and federated learning remain an open problem \cite{selvaraju2017grad}.

\subsection{Privacy and Security Issues}

Although deep neural network models brought great advancement in the medical field, and federated learning prevents a model from private information leakage, various privacy and security attacks remain as unsolved problems. For instance, a medical image deep neural networks are especially susceptible to adversarial attacks due to ambiguous ground truth, highly standardized image, and many other reasons \cite{finlayson2018adversarial}. At the same time, however, the attacks can be easily detected because of the biological characteristic of the images (i.e., manipulation occurring outside the pathological region) \cite{ma2021understanding}. This section will introduce various attack and defense approaches, especially those studied in federated learning environments.

\begin{table*}[]
    \caption{A Summary of Privacy and Security Attacks} 
    \label{tab:attacks}
    \centering
    \begin{tabular}{>{\centering\arraybackslash}m{1.7cm}|>{\centering\arraybackslash}m{2cm}|>{\centering\arraybackslash}m{2.8cm}|>{\centering\arraybackslash}m{3cm}|>{\centering\arraybackslash}m{2.3cm}} \hline\cline{1-5}\cline{1-5}
        \textbf{Attack Category} & \textbf{Attack Types} & \textbf{Attack Target} & \textbf{Attack Methods} & \textbf{Attacker Role}\\ \hline\cline{1-5}
        
        \multirow{6}{*}{\parbox{1.7cm}{\centering{Poisoning Attacks}}}
        & \multirow{2}{*}{\parbox{2cm}{\centering{Data Poisoning}}} & \multirow{2}{*}{\parbox{2.8cm}{\centering{Security \\ (Data Integrity)}}} 
        & Label Flipping                    & Client \\ \cline{4-5}
        & & & Backdoor                      & Client \\ \cline{2-5}
        
        & \multirow{3}{*}{\parbox{2cm}{\centering {Model Poisoning}}} & \multirow{3}{*}{\parbox{2.8cm}{\centering {Security \\ (Model Integrity)}}}  
        & Gradient Manipulation             & Client \\ \cline{4-5}
        & & & Training Rule Manipulation    & Client \\ \hline

        \multirow{6}{*}{\parbox{1.7cm}{\centering Inference Attacks}}
        & \multirow{2}{*}{\parbox{2cm}{\centering {Membership Inference}}} & \multirow{2}{*}{\parbox{2.8cm}{\centering {Privacy \\ (Information Leak)}}}  
        & Membership Inference              & Client \& Server \\ \cline{4-5}
        & & & Properties Inference          & Client \& Server \\ \cline{2-5}
        
        & \multirow{3}{*}{\parbox{2cm}{\centering {GAN Reconstruction}}} & \multirow{3}{*}{\parbox{2.8cm}{\centering {Privacy \\ (Information Leak)}}}  
        & Class Representative Inference    & Client \& Server \\ \cline{4-5}
        & & & Inputs and Labels Inference   & Client \& Server \\ \hline\cline{1-5}\cline{1-5}
    \end{tabular}
\end{table*}

\subsubsection{Attacks}

Federated learning is especially vulnerable to adversarial attacks due to the absence of raw data inspection and the collaborative training using private local data. As generally known, machine learning can be divided into the two phases: training phase and inference phase. Nevertheless, due to zero knowledge distributed nature of federated learning, the training phase attacks are more severe than those of the inference phase; as neither centralized property (i.e., server) nor the other participating clients are allowed to investigate each other's private data.

\subsubsection*{Poisoning Attacks}
Poisoning attacks can be categorized into data poisoning and model poisoning attacks. The two types of poisoning attacks are different in that the former aim to compromise the integrity of the training data, while the latter aim to compromise the integrity of the model. 

Data poisoning attacks include label flipping or data backdoor attacks. Label flipping attacks are one of the client-side data poisoning attacks that flip the labels of the attacker-chosen data classes to attacker-chosen labels so as to miss-classify the specific data classes. 
Tolpegin simulates and analyzes label flipping attacks \cite{tolpegin2020data}, a type of data poisoning attacks. In their experiment, the class label of airplane images is flipped to bird, so the global model misclassifies airplane images to bird at inference time. 
Hayes introduces a contamination attack that is essentially manipulating a small set of training data \cite{hayes2019contamination}, compromising the integrity of the data. The author suggested adversarial training as a defense, which will be discussed in Section 3.4.2 in detail.

The model poisoning attacks involve model backdoors and gradient and/or training rules manipulation. Although poisoning attacks can be differently categorized into two, model poisoning attacks generally include data poisoning attacks as the poisoned data ultimately leads the model to be poisoned. Therefore, we here introduce numerous previous works that are not limited to data poisoning but the hybrid approach of data and model poisoning attacks as well.

Bagdasaryan proposed model replacement to introduce a backdoor into the global model \cite{bagdasaryan2020backdoor}. Their proposed attack kept high accuracy for both main and backdoor tasks so as to improve its persistence by evading anomaly detection. 
Fang manipulated the local model parameters before sending them to the global server \cite{fang2020local}. As a result of the manipulation, the local models deviate towards the inverse direction of the global model before the attack.
Bhagoji introduced a targeted model poisoning attack that poisons the model updates by explicit boosting and remains stealthy by alternating minimization \cite{bhagoji2019analyzing}. 
Xie proposed a distributed backdoor attack that breaks down a global trigger pattern into distinct local patterns and embeds them in the training sets of several adversarial parties \cite{xie2019dba}. Their work showed that the distributed attack is more effective than the centralized backdoor attacks.
Fung pointed out the vulnerabilities of federated learning, especially against the Sybil-based poisoning attacks \cite{fung2018mitigating}. The authors mentioned that the distributed nature increases attack effectiveness, especially when multiple malicious parties participate.

\subsubsection*{Inference Attacks}
Unlike poisoning attacks, inference attacks typically hamper the privacy of private information. Inference attacks include membership inference and GAN-based reconstruction attacks that lead the system to leak information about the training data unintentionally. The recent trend of inference attacks is moving toward the GAN-based method due to its stealth and detection evasion ability. 
Wang achieves user-level privacy leakage through incorporating GAN with a multi-task discriminator \cite{wang2019beyond}. Their proposed method discriminates category, reality, and client identity of input data samples and recovers the user-specific private data. 
In \cite{zhang2019poisoning}, an attacker first acts as a benign participant and stealthy trains a GAN to mimic the other participants’ training samples. With the generated samples, the attacker manipulates the model update with a scaled poisoning model update so as to compromise the global model ultimately.

\subsubsection{Defense Methods}

The defense mechanism against the attacks in federated learning includes minimizing the influence of malicious clients and preventing the malicious clients’ model parameters from incorporating into the global model. Also, with respect to privacy, the defense includes preventing private information from being leaked. 

\subsubsection*{Information Leak Prevention}
The famous technique to prevent the model from leaking private information, Multi-Party Computation (MPC), Homomorphic Encryption, and Differential Privacy (DP), has been widely used. 
PEFL solved the vulnerability by leveraging MPC \cite{hao2019efficient}. Their proposed method has strength in the situation that multiple clients collude so as to prevent private data from being leaked.

The following previous works leverage a combination of those mentioned above three popular techniques. 
Augenstein demonstrated generative models that are trained using federated methods with differential privacy. They applied their model to both text and image, using differentially private federated GANs \cite{augenstein2019generative}.
Ghazi also exploits differential privacy for secure aggregation via shuffled model \cite{ghazi2019scalable}. Their proposed methods preserve privacy and relax computational complexity, which is one of the necessities for better scalability.
Hao combined differential privacy and additive homomorphic encryption to obtain both performance and security \cite{hao2019towards}. Their proposed method is especially robust when not only the clients but the server is honest but curious.  
Truex approached similarly, combined differential privacy with multiparty computation balances the trade-off between the performance and availability \cite{truex2019hybrid}.

\subsubsection*{Model Protection}

In the perspective of security, various works have been done to prevent model from corruption. There are broadly two approaches: robust aggregation and anomaly detection. Typically, robust aggregation aims to train the global model robustly, even if malicious clients participate in the federated learning process. Anomaly detection aims to detect malicious or anomalous clients so that their model updates are not aggregated in the server and/or block them from further participation.

Fu offered a robust aggregation technique with residual-based reweighting \cite{fu2019attack}. Their reweighting strategy employed iteratively reweighted least squares to integrate repeated median regression.
While, various works defended against various attacks by identifying and classifying anomalies \cite{li2020learning, shen2016auror, fang2020local, tolpegin2020data, jeong2021abcfl}. 
Li proposed spectral anomaly detection mechanism based on models’ low-dimensional embeddings \cite{li2020learning}. The central server learns to detect and remove the malicious model updates by removing noisy features and retaining essential features, leading to targeted defense. A notable point of their proposed approach is that it worked in both semi-supervised or unsupervised learning and designed the protocol for encryption.
FoolsGold mitigated poisoning attacks \cite{wu2020mitigating}. Their approach comes from the idea that the malicious clients' updates are different from those of the benign ones; thereby, the Sybils are distinguishable by measuring the contribution similarity.

The aforementioned work's objective, anomaly detection, remains the same, but several approaches leveraged clustering-based and thresholding-based approaches \cite{shen2016auror, fang2020local, tolpegin2020data, jeong2021abcfl, cao2020fltrust, sun2019can}. 
Auror dealt with targeted poisoning attacks leveraging clustering and thresholding techniques. Shen created clusters of clients and measured the pairwise distance between them, which will be used to distinguish two distinctive clusters. Within each cluster, if more than half of the clients are determined to be malicious by the predefined threshold, all the clients belonging to that cluster are then classified as malicious \cite{shen2016auror}. 
Fang proposed a defense mechanism, a combination of error rate-based rejection and loss function-based rejection \cite{fang2020local}. The idea comes from that as the malicious clients tamper with the models' performance, the error rate and the loss impact are greater than those of benign clients. Therefore, if a client greatly impacts the higher error and loss rate, the clients are identified as malicious so as to aggregate only benign clients' weights. 
Along with the data poisoning attack, Tolpegin used PCA to visualize the spatially separable malicious clients' model updates and those of benign ones \cite{tolpegin2020data}. 

Sun defended against backdoor attacks by leveraging norm bounding and weak differential privacy \cite{sun2019can}. They noted that the norm values of malicious clients are relatively greater than those of benign ones, so they detected malicious clients by thresholding based on the calculated norm value of each clients' weight updates. 
FLTrust protected the global model against byzantine attacks by making use of a ReLU-clipped cosine similarity-based trust score \cite{cao2020fltrust}. In their works, however, the global server had been trained on an innocent dataset called root dataset; in other words, the server did have the knowledge of benignity and malice of the client dataset, which is a breach of the no-raw-data-sharing assumption.
Based on \cite{cao2020fltrust, sun2019can}, Jeong \cite{jeong2021abcfl} proposed ABC-FL that leverages feature dimension reduction, dynamic clustering, and cosine-similarity-based clipping to detect and classify anomalous and benign clients where the benign ones have non-IID data and IID data. Similar to the aforementioned approaches, only the benign clients' model weights are aggregated in the global server.

\section{Modular Framework Under Development}

Although there has been an increasing number of studies in the medical field using federated learning, a variety of open problems still remain. An easier and more convenient federated learning framework is demanding as this field of study getting more attention. 

Our research team is developing a lightweight framework, \textit{Modular FL}, that modularizes the essential component of federated learning. The Modular FL consists of the federated learning core component, serving component, and communication channel.

\begin{figure}[h]
\centering
\includegraphics[width=0.95\textwidth]{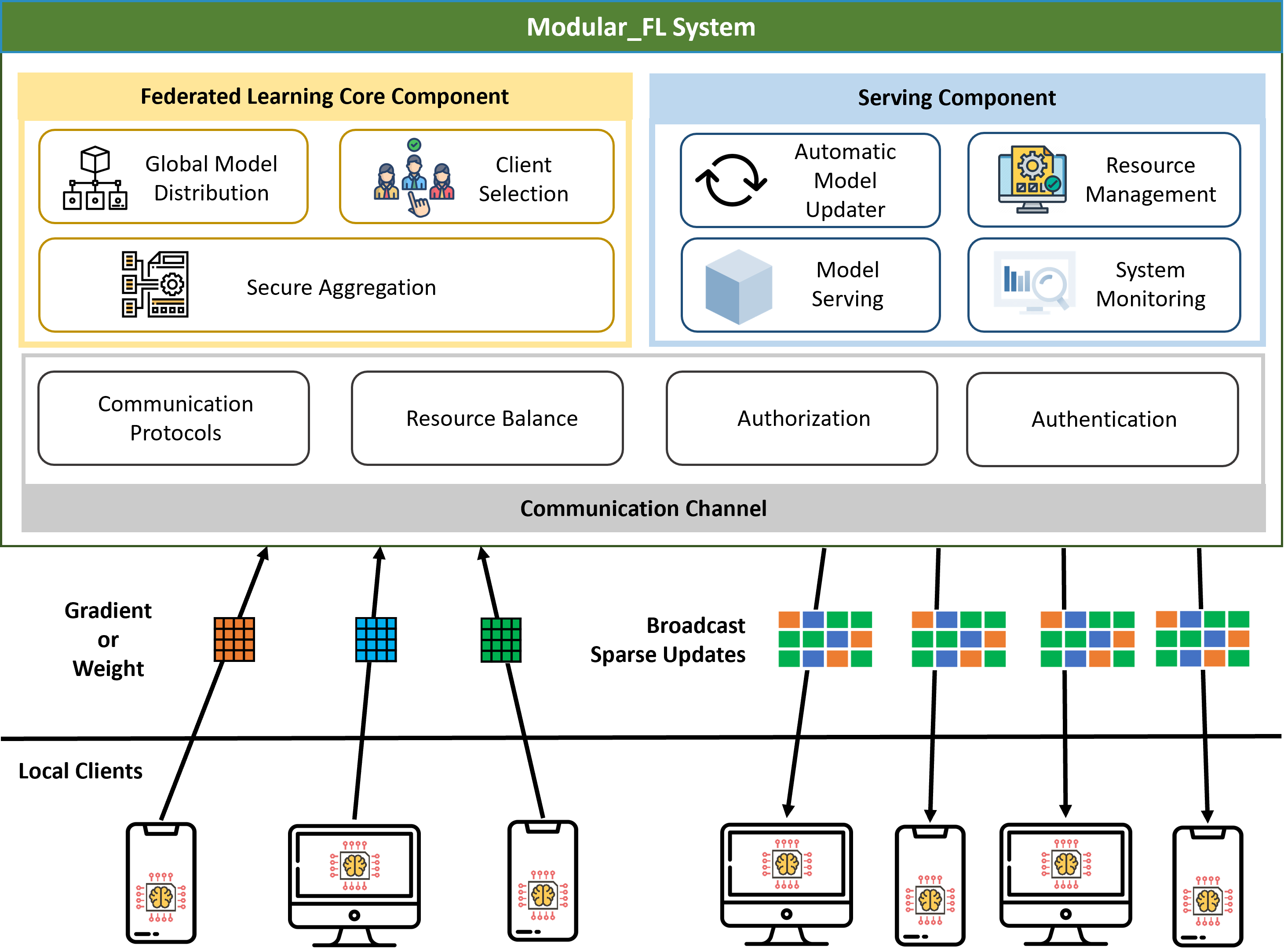}
\caption{The architecture of the Modular FL currently in development.}
\label{fig: modularfl}
\end{figure}

A federated learning core component controls the core steps of federated learning between a server and multiple clients. This component consists of four modules, namely global model distribution, client selection, and secure aggregation modules. 
Global model distribution module is in charge of initializing and distributing a global model to the client selected by client selection module. A secure aggregation module is responsible for aggregating the weight vectors received as a result of each local training. 
A serving component also consists of four modules, automatic model updater, model serving, resource management, and system monitoring, providing management related functionalities. We omit the explanation of each serving component module, as their names are self-explanatory. 
The server also manages communication protocols, balances shared resources, authorizes the clients, and authenticates the clients. The securely aggregated global model is advertised to local clients through communication channel and with the aid of the server's management.

Each component includes the learning and communication functions, providing researchers a structure to experiment by simply changing the functions to address the remaining issues. For example, researchers can change the client selection function to add an incentive mechanism to compensate \textit{good quality} clients.  
Modular FL is also easily expandable due to its flexible structure, where model weights from learning process can be saved as Json. This characteristic leads Modular FL being not limited to a library or even a language. 

\section{Conclusion and Future Works}

Federated learning, which has emerged under strong data protection, has been extensively applied in various research fields. Various researches are being conducted to apply to medical fields that require data privacy, especially in radiology, pathology, and neuroscience. The application of federated learning in the medical field, however, raises many different research issues.

In this survey, we review the issues of federated learning, especially those that can occur in the medical field. We summarize the various works on different issues, so our work may be a useful resource for researchers studying federated learning. Even though various attempts have been made to address many issues that can possibly happen in the medical field, there still remain open questions. We provide some example open questions and future direction in the following. \\

\noindent\textbf{Explainable AI and Federated Learning.} Not only in the federated learning, explainability has been an important issue in machine learning due to its black-box nature. In federated learning, specifically when the raw data cannot be investigated, gaining explainability of the result is especially difficult. Even though there have been some work leveraging explainable AI in federated learning, it has not yet been extensively studied. For the more accurate and fail-safe usage of federated learning, more works have to be equipped with explianability. \\

\noindent\textbf{Data Heterogeneity, Attack, and Federated Learning.} Even though numerous works on various attacks and defense mechanisms have been published, attacks with heterogeneous data has not yet been exhaustively studied. Nevertheless, the real-world data can be non-IID, attacks and defenses with those data distribution must be addressed. Since federated learning will be widely used to preserve data privacy and reduce communication complexity, researchers should study attacks and defenses even when some clients have highly non-IID data. 
\\

\section*{Author Contribution}
J.H.Y. and H.J. are co-first authors and contributed equally to the composition and preparation of the paper. T-M.C. proposed research and theorem in the field of federated learning in medical application. J.H.Y., H.J., J.L., and T-M.C. discussed about suggested research idea. J.H.Y. and H.J. wrote the manuscript with support from J.L.. H.J. revised the entire manuscript. T-M.C. provided invaluable guidance throughout the research and the writing of the manuscript. All authors have reviewed the manuscript.

\section*{Acknowledgement}
This research was supported by Institute of Information \& communications Technology Planning \& Evaluation (IITP) grant funded by the Korea government(MSIT) (No.2020-0-00990, Platform Development and Proof of High Trust \& Low Latency Processing for Heterogeneous·Atypical·Large Scaled Data in 5G-IoT Environment)

\bibliographystyle{splncs04}
\bibliography{biblio}

\end{document}